\documentclass[11pt,a4paper]{article}
\usepackage[hyperref]{emnlp2020}
\usepackage{times}
\usepackage{latexsym}

\usepackage{graphics}
\usepackage{graphicx}
\usepackage{xcolor, soul}
\usepackage{marvosym}
\usepackage{amssymb}
\usepackage{microtype}

\aclfinalcopy 

\title{Leveraging Event Specific and
Chunk Span features to Extract COVID Events from tweets}

\author{Ayush Kaushal
 \and Tejas Vaidhya
\\
  Indian Institute of Technology, Kharagpur \\
  \texttt{ ayushk4@gmail.com, iamtejasvaidhya@gmail.com} \\}

\date{}

\begin{document}
\maketitle
\begin{abstract}

Twitter has acted as an important source of information during disasters and pandemic, especially during the times of COVID-19.
In this paper, we describe our system entry for \textit{WNUT 2020 Shared Task-3}. The task was aimed at automating the extraction of a variety of COVID-19 related events from Twitter, such as individuals who recently contracted the virus, someone with symptoms who were denied testing and believed remedies against the infection.
The system consists of separate multi-task models for slot-filling subtasks and sentence-classification subtasks while leveraging the useful sentence-level information for the corresponding event. The system uses COVID-Twitter-Bert with attention-weighted pooling of candidate slot-chunk features to capture the useful information chunks.
The system \textbf{ranks 1st} at the leader-board with \textbf{F1 of 0.6598}, without using any ensembles or additional datasets.
The code and trained models are available at this https url\footnote{https://github.com/Ayushk4/extract\_covid\_entity}.

\end{abstract}
\section{Introduction}

The World Health Organization declared COVID-19, a global pandemic on March 11, 2020. As of 2020/09/21, there are over 30 million cases\footnote{https://coronavirus.jhu.edu/map.html} and 900,000 deaths due to the infection. With the imposed lockdown, work from home and physical distancing, social media like twitter saw an increased usage. A large part of the use was posting and consuming information on the novel infection. These information include potential reasons for contraction of the disease, such as via exposure to a family member who tested positive, or someone who is showing COVID symptoms but was denied testing. Accompanying to the pandemic was an infodemic of misinformation about COVID-19, including fake remedies, treatments and prevention-suggestions in social media \cite{covid_infodemic}.

\citet{zong2020extracting} show the possibility to automatically extract structured knowledge on COVID-19 events from Twitter and released a dataset of COVID related tweets across 5 event types. We used this dataset in our experiments for the shared-task. These tweets are annotated for whether they belong to an event (we refer to this as the \textbf{event-prediction} task in this paper) and their event-specific questions (factual or opinion). We identify these event-specific questions into two types of subtasks,  \textbf{slot-filling} and \textbf{sentence classification}.

Our system consists of separate multi-task models for slot-filling subtasks and sentence-classification subtasks.
Our contribution comprises improvement upon the baseline
(mentioned in section \ref{sec:rl}) in three ways:

\begin{itemize}
    \item We incorporate the event-prediction task as auxiliary subtask and fuse its features for all the event-specific subtasks.
    
    \item We perform an attention-weighted pooling over the candidate chunk span enabling the model to attend to subtask specific cues.
    
    \item We use the domain-specific Bert of Covid-Twitter Bert \cite{ctbert}.
\end{itemize}

\section{Related Works}\label{sec:rl}

Sentence classification tasks (such as opinion or sentiment mining) as well as slot-filling tasks have greatly progressed with deep learning advancements such as LSTM \cite{lstmpaper}, Tree-LSTM \cite{treelstmpaper} and transfer learning over pre-trained models \citep{elmopaper, ulmfitpaper, bertpaper}. 
Among these, CT-Bert outperforms others on COVID related twitter tasks \citep{ctbert}. Taking inspiration from the same, we use CT-Bert as part of our architecture.
A variety of slot-filling approaches have been built on top of these deep learning advancements \citep{slot_filling1, slot_filling2}. The proposed baseline for our task \cite{zong2020extracting} modifies Bert model for slot-filling problem inspired by \citet{baldinisoares-etal-2019matching}. Due to the excellent performance offered by Bert \cite{bertpaper} and \citet{baldinisoares-etal-2019matching}, we build upon this baseline approach.

Extraction of structured knowledge from tweets pertaining of events \cite{benson_etal_2011_event} has been studied for disaster and crises management \cite{disaster_events1, disaster_events3} and in pandemic scenarios \cite{social_media_for_pandemic}. Extracting such entities can be useful for epidemiologists, deciding policies and preventing spread \cite{social_media_for_pandemic, zong2020extracting}.

Due to the fast-spreading nature of the infection, it is also difficult to manually trace the spread of the pandemic. However, with twitter event-specific entity extraction and Geo-location, one could potentially build a real-time pandemic surveillance system \citep{surveillance1, surveillance2}. \citet{Bal_Sinha_Dutta_Joshi_Ghosh_Dutt_2020} show that health-issues related misinformation is prevalent in social media, while \citet{covid_infodemic} talks about covid-specific misinformation. Such systems for extracting structured knowledge over the tweets talking about potential cures for COVID will help study how users perceive the COVID misinformation.

In \S\ref{problem_section}, we describe the dataset and the problem statement. Then in \S\ref{approach_section}, we discuss the details of our two multi-task models followed by experiments, results and conclusion.

\section{Dataset and Problem statement} \label{problem_section}

Now, we will briefly go over the dataset. The reader may refer \cite{zong2020extracting} for full details. Each of the 7500 tweets in the dataset belongs to one of the 5 event types: tested-positive, tested-negative, can-not-test, death, and cure. The first four events aimed at extracting structured reports of coronavirus related events, such as self-reported cases or news stories about public figures who were exposed to the virus.
Each tweet was first annotated for whether it belongs to its respective event (e.g. Is the tweet belonging to the tested-positive event talking about someone who tested positive?). Throughout this paper, we refer to this as the \textbf{Event-Prediction} task. The tweets that correspond to its event were then annotated for event-specific questions or \textbf{subtasks} about factual information and user’s opinions. All annotations are done by multiple Amazon Mechanical Turks with inter-annotation agreement. The event-specific questions or subtasks (e.g. name, age, gender of the person tested positive) varies depending on the event. These subtasks are of two categories: \textbf{slot-filling} (e.g., Who tested positive/negative?, Where are they located?, Who is in close contact with person contracting the disease?) and \textbf{sentence classification} (e.g. Is author related to infected person?, Does the author experience any symptoms?, Does the author believe a cure method is effective?).

\begin{figure}
    \centering
    \includegraphics[scale=0.28]{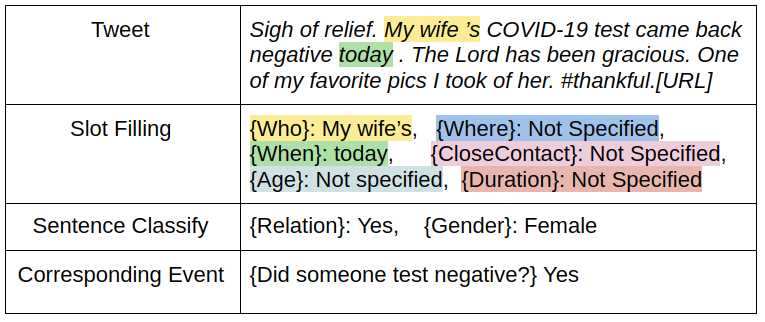}
    \caption{An example tweet from tested negative event.}
    \label{fig-dataset_example}
\end{figure}

The dataset released tweet IDs and their annotations. We obtain our text corresponding to tweets using the official Twitter API\footnote{https://developer.twitter.com/}. Table \ref{Table1:} shows the statistics for the dataset we scrapped in early July.\footnote{We get about 350 fewer tweets than the corpus. Some tweets are not obtainable over time as the accounts/tweets get deleted, renamed, banned, or change-visibility etc.} Figure
\ref{fig-dataset_example} shows an annotated example from the dataset. We identify the event-specific subtasks into two categories shown in Table \ref{table2}.

We now formally describe the two types of event-specific subtasks:

\begin{table}
 \begin{center}
  \begin{tabular}{c | c } 
   \hline
   \textbf{Event} & \textbf{\# Tweets} \\ [0.5ex] 
   \hline
   Tested positive & 2397 \\ 
   Tested negative & 1144 \\
   Can Not Test & 1128 \\
   Death & 1231 \\
   Cure/Prevention & 1244 \\
   \hline
   Total & 7144 \\
   \hline
  \end{tabular}
  \caption{\label{Table1:} Dataset statistics, scraped during early July.
  }
 \end{center}
\end{table}

\begin{table*}[!htb]
  \centering

   \begin{tabular}{p{2.4cm}|p{3.80cm}| p{8.50cm}}

   \textbf{Event} & \textbf{Sentence Classification} & \textbf{Slot-Filling task} \\
   \hline
    Tested positive & gender, relation & who,age,recent-visit,when,where,employer,c.-contact \\
    Tested negative & gender, relation & who,age,when,where,duration,close-contact\\
    Can Not Test & relation, symptoms & who,when,where \\
    Death & relation, symptoms & who,age,when,where\\
    Cure & opinion & what is the cure, who is promoting cure \\

  \end{tabular}
  \caption{The proposed event-specific subtasks split into two subtask types: slot-filling and sentence classification}
  \label{table2}
\end{table*}

\textbf{Slot-filling subtasks}: Assume $n$ slot-filling subtasks $\{S_1, S_2 ... S_n\}$. We set up each slot-filling subtask $S_i$ as a supervised binary classification problem. Given the tweet $t$ and the candidate slot $s$, the model $f(t, s) \to \{0, 1\}$ predicts whether $s$ answers its designated question. We extract a list of candidate slot of all noun chunks and name entities in each of the tweets by using a Twitter tagging tool \cite{ritteretal-2011named} same as the baseline.

\textbf{Sentence classification subtasks}: Assume $m$ sentence classification subtasks $\{C_1, C_2 ... C_m,\}$. Given a sentence classification subtask $C_i$ aims to learn a model $g(t) \to \{l_1, l_2 ... l_{k}\}$, where $t$ is a tweet and $l_j$ is a label. Here the number of labels can vary depending on the subtask, for example, gender is labelled with \{Male, Female, Others/Not Specified\}, Relation with \{Yes, No\}, Opinion with \{effective, no cure, not effective, no opinion\} and so on. All these subtasks are `supervised' classification problems.

The dataset is also annotated with whether a tweet corresponds to its respective event or not. We treat this as an additional \textbf{Event-Prediction task}. This is a binary classification task that aims to learn a model $h(t) \to {0, 1}$ where $t$ is a tweet.


    
    

\section{Approach}\label{approach_section}


In the following subsections \S \ref{slot_filling_subsection} and \S \ref{sentence_classification_subsection}, we describe our multi-task model for slot-filling and sentence-classification respectively.

\subsection{Slot-filling} \label{slot_filling_subsection}
\begin{figure}
\centering
  \includegraphics[scale=0.9]{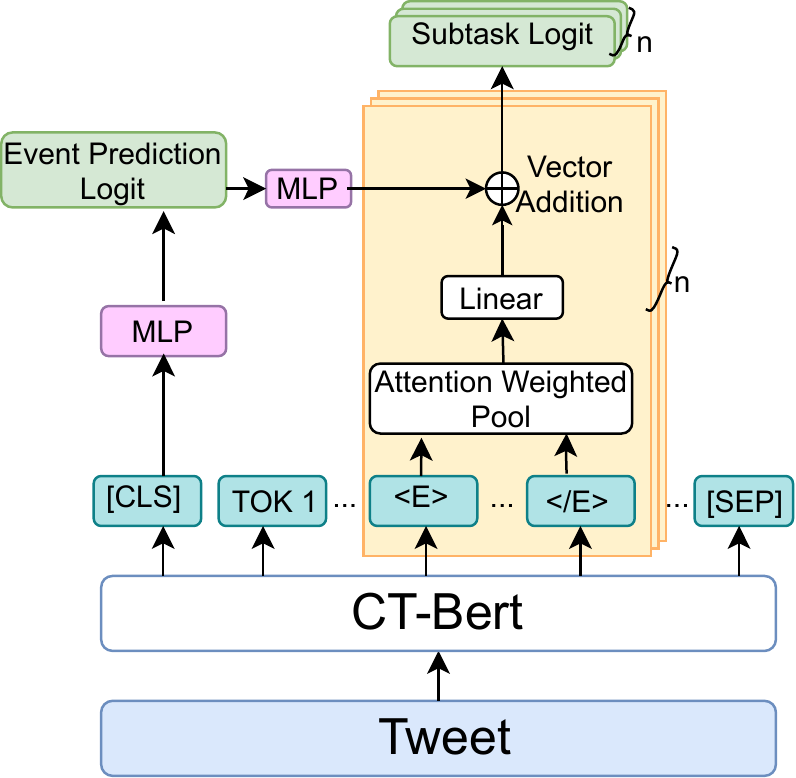}
    \caption{Slot-Filling Model, described in Section \S\ref{slot_filling_subsection}. Here n is the number of slot-filling subtasks. }
    \label{fig:model_slot}
\end{figure}

We improve upon the baseline \cite{zong2020extracting} by using domain-specific Bert, using attention-weighted pooling over the candidate chunk feature sequence, incorporating auxiliary Event-Prediction task and utilizing its logits for all the slot-filling subtasks. Before describing the approach, we first describe the Bert baseline. Our slot-filling model can be seen in figure \ref{fig:model_slot}.

The baseline consists of Bert based classifier. It takes a tweet $t$ as input and encloses the candidate slot $s$, within the tweet, inside special entity start $< E>$ and end $</E>$ markers. The Bert hidden representation of token $<E>$ is then processed through a fully connected layer with softmax activation to make the binary prediction for a task \cite{baldinisoares-etal-2019matching}. Since many slot-filling tasks within an event are semantically related to each other, they jointly trained the final softmax layers of all the subtasks $S_i$ in an event by sharing their Bert model parameters.

COVID Twitter Bert (CT-Bert) is a Bert-Large model pretrained on Twitter Corpus on COVID-19 topics, leading to marginal improvements from Bert on tasks based on Twitter datasets\cite{ctbert}. This motivates us to use CT-Bert instead of Bert from the baseline model.

The baseline, uses the Bert hidden representation of token $<E>$ for classification. Here, however, we use attention-weighted pool of the CT-Bert hidden representation of tokens between $<E>$ and $</E>$ (both inclusive). Formally, let $\{x_0, ... x_p, ... x_q, ...x_n \}$ be the output vectors from the hidden representation of CT-Bert where $p$ and $q$ are indices of $<E>$ and $</E>$ respectively, then for any of the slot-filling subtask $S_j$, we get its pooled vector as follows:

\begin{equation}
\widetilde{x}^{S_j} = \sum_{i=p}^{q}\alpha_i^{S_j}x_i \end{equation}
$$\alpha_i^{S_j} = Softmax_{p\ to\ q}(x_i^T a^{S_j})$$

where ${x_i}^T$ denotes the transpose of $x_i$, $a^{S_j}$ is a trainable vector. The motivation for attention weighted pooling is that depending on the task, model can attend to different portions of the candidate slot chunk. Next we obtain the binary classification score vector:

\begin{equation}
\displaystyle{h^{S_j} = W^{S_j}\widetilde{x}^{S_j} + b^{S_j}}
\label{eq-1}
\end{equation}

Here $W^{S_j}$ and $b^{S_j}$ are trainable parameters.

We treat the Event-Prediction task as an auxiliary task and then fuse its logits to each of the other slot-filling subtasks. The motivation is that a task-specific entity shall be present in a tweet only if the tweet belongs to its respective event.

To predict the label for Event-Prediction task, we take the CT-Bert features of $[CLS]$ token and pass it through a MultiLayer Perceptron (MLP) to get logits $h_{ces}$.

We fuse $h_{ces}$ prediction over each subtasks ${S_j}$ by adding it to $h^{S_j}$ (from (\ref{eq-1})) to get the logits $h_f^{S_j}$:

\begin{equation}
h_f^{S_j} = h^{S_j} + MLP^{S_j}(h_{ces})
\end{equation}
In practice, we share the parameters of the $MLP^{S_j}$ across all the slot-filling subtasks $S_j$.

Given a tweet $t$ and slot $s$, our loss for slot-filling model over $n$ slot-filling subtasks $\{S_1, S_2 ... S_n\}$ and Event-Prediction task looks like:

$$Loss(t, s, y_{ces}, (y_1, y_2 ... y_n))$$
\begin{equation}
= \lambda_1 CE_{Loss}(h_{ces}, y_{ces}) + \sum_{k=1}^{n}CE_{Loss}(h_f^{S_k}, y_k)
\end{equation}

where $CE_{loss}$ is softmax cross entropy loss, $y_{ces}$ is ground truth label for Event-Prediction task and $(y_1, y_2 ... y_n)$ are the labels for the candidate slot $s$ of tweet $t$ for the subtasks $\{S_1, S_2 ... S_n\}$. We keep $\lambda_1$ = 1.

Our preprocessing for this is same as baseline.

\subsection{Sentence classification}\label{sentence_classification_subsection}

\begin{figure}
\centering
  \includegraphics[scale=1.05]{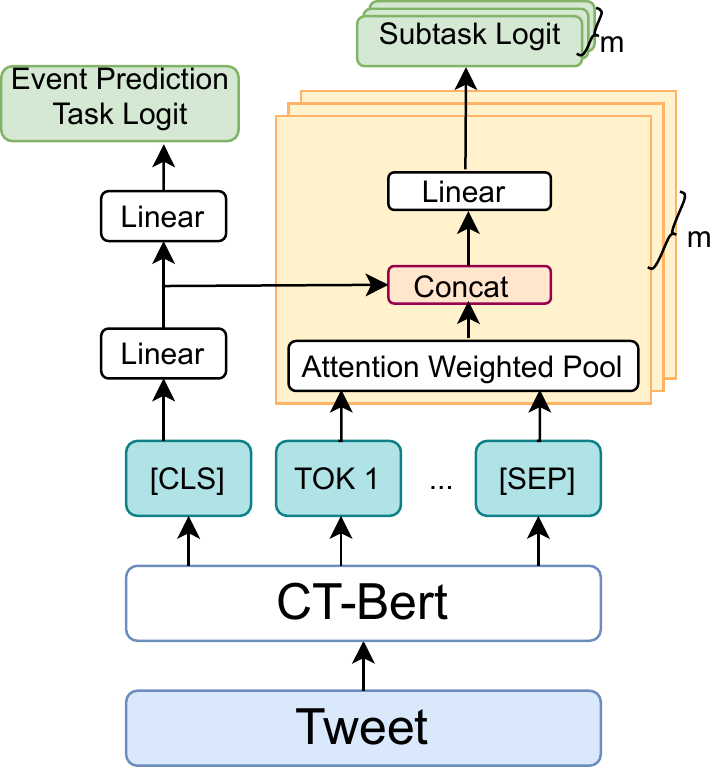}
    \caption{Sentence Classification model, described in section. \S\ref{sentence_classification_subsection}. Here m is the number of Sentence Classification subtasks. }
    \label{fig:model_sent_class}
\end{figure}

Our Sentence classification model is shown in figure \ref{fig:model_sent_class}. We use a Bert based sentence classifier and improve it by using CT-Bert, incorporating the auxiliary Event-Prediction task and attention-weighted pooling over the entire sequence.

This model uses CT-Bert instead of Bert and the auxiliary Event-Prediction task for same reason as the slot-filling model.
 
An attention-weighted pooling is done over the feature sequences from CT-Bert to extract the most relevant information. Formally, let $\{x_0, x_1,... ...x_n \}$ be the output vectors from CT-Bert (here $0$ and $n$ are indices of $[CLS]$ and $[SEP]$ respectively). Then for any of the sentence classification subtask $C_j$, we get its pooled vector $\widetilde{x}^{C_j}$ as follows:

\begin{equation}
\widetilde{x}^{C_j} = \sum_{i=0}^{n}\beta_i^{C_j}x_i
\label{equation:sent_1}
\end{equation}
$$\beta_i^{C_j} = Softmax_{i}(x_i^T a^{C_j} + c^{C_j})$$

where $a^{C_j}$, $c^{C_j}$ are trainable vector and scalar respectively.

For the Event-Prediction task, we take the CT-Bert vector representation of $[CLS]$ token and pass it through a MLP. Assume the MLP's final and hidden states to be $v_{ces}$ and $h'_{ces}$.

Next, we incorporate information from Event-Prediction task into sentence classification subtask $C_j$. Since the sentence classification subtasks aren't binary classification, so, unlike the slot-filling model, we cannot merely add the Event-Prediction logits to all tasks. Additionally, we desire sentence-level event specific features for each of the sentence level predictions. Hence, we concatenate the hidden state features from the MLP of Event-Prediction task $h'_{ces}$ to pooled vector $\widetilde{x}^{C_j}$ from \ref{equation:sent_1} to get the logits $h_f^{C_j}$ for each subtask ${C_j}$, as follows:

\begin{equation}
\displaystyle{h_f^{C_j}} = [\widetilde{x}^{C_j} ; h'_{ces}]^T W^{C_j} + b^{C_j}
\end{equation}

Here ${ }^T$ denotes transpose, $[ ; ]$ denotes vector concatenation. $W^{C_j}$ and $b^{C_j}$ are trainable.

Given a tweet $t$, our loss for sentence classification model over $m$ sentence classification subtasks $\{C_1, C_2 ... C_m\}$ and Event-Prediction task is:

$$Loss(t, y_{ces}, (y_1, y_2 ... y_m))$$
\begin{equation}
= \lambda_2 CE_{Loss}(v_{ces}, y_{ces}) + \sum_{k=1}^{m}CE_{Loss}(h_f^{C_k}, y_k)
\end{equation}

where $CE_{Loss}$ is softmax cross entropy loss,  $y_{ces}$ is ground truth label for Event-Prediction task and $(y_1, y_2 ... y_m)$ are the labels for tweet $t$ for the subtasks $\{C_1, C_2 ... C_m\}$. We keep $\lambda_2$ = 1.

Preprocessing for sentence classification is done using ekphrasis library \cite{baziotis-pelekis-doulkeridis:2017:SemEval2}. We remove Emoji, URL, Email, punctuation and normalize text by word segmenting, lower-casing and word decontraction.

\section{Experiments}\label{experiments_section}

All the experiments were performed using PyTorch \cite{pytorch} and Hugging Face's transformers \cite{huggingface}. We use git and wandb \cite{wandb} for experiment tracking. Optimization is done using Adam \cite{kingma2014adam} with a learning rate of 2e-5. Slot-filling models are trained for 8 epochs and sentence classification model for 10 epochs. Average training time per epoch on Tesla P100 is $\approx$ 4 minutes for slot-filling, and $\approx$ 30 second for sentence classification.

We use a 70-30 split for train-valid set. The valid set is used to obtain the best threshold for each of the slot classification tasks over the grid $\{0.1, 0.2, ..., 0.9\}$. We exclude labels with ``No consensus" from our data.\footnote{As per the submission guidelines, some subtasks like opinion had their label classes merged. We incorporate these changes in our model.}

All the MLP have 1 hidden layer and 0.1 dropout. $MLP_{S_j}$ has 4 hidden size, LeakyReLU activation \cite{Maas13rectifiernonlinearities} with 0.1 negative slope, rest of the MLP have 50 hidden size and Tanh activation.

\section{Results}

Our performance on the held-out test set is shown in Table \ref{Table3:}. Our system \textbf{ranks 1st position} in the W-NUT 2020 Shared Task-3 \cite{zong2020extracting}. We also independently rank 1st for 3 of the 5 events: `Can Not Test', `Death', and `Cure'.


\begin{table}
 \begin{center}
  \begin{tabular}{p{2.5cm}|p{0.5cm}p{0.5cm}p{0.5cm}}
   \hline
   \textbf{Event} & \textbf{F1} & \textbf{P} & \textbf{R}\\ 
   \hline
   Tested Positive & .68 & .80 & .58 \\
   Tested Negative & .66 & .66 & .67 \\
   Can Not Test  & .65 & .67 & .64 \\
   Death           & .69 & .72 & .67 \\
   Cure/Prevention & .63 & .75 & .53 \\
   \hline
   \hline
   \textbf{Overall}         & \textbf{.66} & \textbf{.73} & \textbf{.60} \\
   \hline
  \end{tabular}
  \caption{\label{Table3:} Micro averaged scores on the held out test set for our final submission.}
 \end{center}
\end{table}

Now we discuss our various experiments.

\paragraph{Slot-filling:}

We experimented with a variety of architectures for slot-filling model. \textbf{Our (SF)} is our Slot-Filling Model from \S\ref{slot_filling_subsection}. \textbf{Our (SF) w/o pool} is our slot-filling model that uses the CT-Bert hidden representation of token $<E>$ to classify instead of doing an attention-weighted pooling. \textbf{Our (SF) w/o CES} is our slot-filling model without Event-Prediction task. \textbf{CT-Bert} and \textbf{Bert-large} are baseline models using CT-Bert and Bert-large instead of Bert-base.

Table \ref{Table4:} shows the performance of these models. There is a considerable performance difference by using CT-Bert instead of Bert, demonstrate the benefits of domain specific pre-training.
\textit{Our (SF) w/o pool} and \textit{Our (SF) w/o CES} outperform CT-Bert demonstrating the importance of Event-Prediction task and  attention-weighted pooling over slot-chunk respectively.
\textit{Our (SF)} using CT-Bert with Event-Prediction and attention-weighted pooling performs the best among these models.

\begin{table}
 \begin{center}

  \begin{tabular}{p{3cm}p{1.5cm}p{1.5cm}}
    \hline
    Model & Micro F1 & Macro F1\\
    \hline
   Our (SF)          & \textbf{.684} & \textbf{.558} \\
   Our (SF) w/o pool & .678          & .557 \\
   Our (SF) w/o CES  & .665          & .552 \\
   CT-Bert           & .662          & .551 \\
   Bert (large)      & .610          & .529 \\ 
   Bert (baseline)   & .612          & .528 \\ 
   \hline

  \end{tabular}
  \caption{\label{Table4:} Results of slot-filling models on our 70-30 split. We report results on the valid set across \textit{all slot filling subtasks} across the 5 events.}
 \end{center}
\end{table}

\paragraph{Sentence level tasks:}

We experimented with various architectures for sentence level tasks. \textbf{Our (SC)} is our Sentence Classification architecture from \S\ref{sentence_classification_subsection}. \textbf{Our (SC) w/o CES} is our Sentence Classification without Event-Prediction task. \textbf{Bert multitask} model predicts using the $[CLS]$ representation from Bert \cite{bertpaper}. We also build an LSTM model \cite{lstmpaper} with GloVe embedding \cite{pennington2014glove}, and twitter-tokenization using WordTokenizers package \cite{worktokenizers}.

Table \ref{Table5:} shows the performance of these architectures. Our (SC) outperforms others on macro F1 and micro F1, followed by Our (SC) w/o CES. The performance difference between these two, shows the benefits of including the Event-Prediction task. While the performance difference between CT-Bert multitask and Our (SC) w/o CES shows the gains from attention weighted pooling. CT-Bert also outperforms Bert multitask, showing its usefulness in our proposed system over using Bert. Lastly, Bert multitask, and all the models using Bert/CT-Bert outperform LSTM by a very large margin demonstrating the superiority of these pretrained language models.

\begin{table}
 \begin{center}
  \begin{tabular}{p{3cm}p{1.5cm}p{1.5cm}}
    \hline
    Model & Micro F1 & Macro F1\\ 
    \hline
    Our (SC) & \textbf{.788} & \textbf{.767} \\
    Our (SC) w/o CES & .777 & .731\\
    CT-Bert multitask & .760 & .717 \\
    Bert multitask & .715 & .612\\
    LSTM multitask & .614 & .543\\
    \hline

  \end{tabular}
  \caption{\label{Table5:} Results sentence classification models on our 70-30 split. We report results on the valid set across \textit{all sentence classification subtasks} across the 5 events.}
 \end{center}
\end{table}

\paragraph{Separate Sentence classification and slot filling models:} Consider \textbf{Bert separate}, a simple system treating the two categories of tasks separately. It has the Bert baseline as its slot filling model and a simple Bert sentence classifier using features from $[CLS]$ for sentence prediction. Bert separate does not have the event-prediction auxilliary task or any attention weighted pooling. Table \ref{Table6:} shows the performance of \textit{Bert separate} against the baseline. \textit{Bert separate} outperforms the Bert baseline by a considerable margin, thus showing the importance of treating the two subtasks differently.

\begin{table}
 \begin{center}
  \begin{tabular}{p{3cm}p{1.5cm}p{1.5cm}}
    \hline
    Model & Micro F1 & Macro F1\\ 
    \hline
    Bert Separate & \textbf{.631} & \textbf{.545}\\
    Bert Baseline & .608 & .512\\
    \hline

  \end{tabular}
  \caption{\label{Table6:} Results comparing the systems treating the sentence classification and slot-filling subtasks separately vs those treating it similarly. We report results on the valid set across \textit{all the subtasks} of both categories across the 5 events.}
 \end{center}
\end{table}

\section{Conclusion and Future Work}

In this paper, we presented our system that bagged 1st position in the WNUT-2020 Shared Task-3 on Extracting COVID Entities from Twitter. We divided the event-specific subtasks into slot-filling and sentence classification subtasks, building separate architectures for the two. For both architectures, we used COVID-Twitter Bert, weighted-attention pooling over chunk-spans/sentence and fused logits and features from auxiliary Event-Prediction task. Our ablation studies demonstrated the usefulness of each component in our system.

There is a lot of scope of improvement for subtasks with few positive labels. Pretraining on relevant data (such as COVID-misinformation datasets for event cure) is a promising direction.

Another direction would be to reduce the training and inference time of slot-filling model by \textit{not} enclosing the candidate chunk within special start $<E>$ and special end $</E>$ tokens. We can instead use the attention-weighted pooling over candidate slot chunks. This will reduce the number of Bert forward passes from $O(k)$ to $O(1)$, where $k$ is the number of candidate chunks in a tweet.

\section*{Acknowledgments}

We are very grateful for the invaluable suggestions given by Nikhil Shah, Dibya Prakash Das and Sayan Sinha. We also thank the organizers of the Shared Task-3 at WNUT, EMNLP-2020.

\bibliography{anthology,emnlp2020}
\bibliographystyle{acl_natbib}

\end{document}